\documentclass[lettersize,journal]{IEEEtran}
\usepackage[T1]{fontenc}

\usepackage{amsmath,amsfonts}
\usepackage{algorithmic}
\usepackage{algorithm}
\usepackage{array}
\usepackage[caption=false,font=normalsize,labelfont=sf,textfont=sf]{subfig}
\usepackage{textcomp}
\usepackage{stfloats}
\usepackage{url}
\usepackage{verbatim}
\usepackage{graphicx}
\usepackage[table,xcdraw]{xcolor}
\usepackage{booktabs}

\definecolor{headerred}{HTML}{B96E64}
\definecolor{rowred}{HTML}{F3E6E4}
\definecolor{headerblue}{HTML}{809FEF}
\definecolor{rowblue}{HTML}{E9EEFB}
\definecolor{headergreen}{HTML}{4FAD6F}
\definecolor{rowgreen}{HTML}{E7F4EC}

\usepackage{hyperref}
\usepackage{cleveref}
\usepackage{soul}
\usepackage{newtxtext}
\usepackage{multirow}
\usepackage{array}
\usepackage{float}
\usepackage{makecell}  
\usepackage[skip=4pt]{caption}  
\usepackage{soul}
\usepackage{xspace}
\usepackage{siunitx}

\usepackage[
  backend=biber,
  style=numeric-comp,
  sorting=none,
  doi=false,
  url=false,
  maxnames=2,
  minnames=1
]{biblatex}

\AtEveryBibitem{
    \clearfield{doi}
    \clearfield{url}
    \clearfield{isbn}
    \clearfield{issn}
    \clearfield{pages}
    \clearfield{publisher}
    \clearfield{address}
    \clearfield{month}
    \clearfield{note}
    \clearfield{volume}
    \clearfield{number}
    \clearfield{series}
    \clearfield{location}
}

\hypersetup{
    pdfauthor={Anonymous},
    pdfproducer={Anonymous},
    pdfcreator={Anonymous}
}

\begin{document}

\title{Transferability Through Cooperative Competitions}
\author{Rodrigo Serra$^{1}$, Carlos Azevedo$^{2}$, André Silva$^{1}$, Kevin Alcedo$^{1}$, Quentin Rouxel$^{3}$, Peter So$^{4}$, Alejandro Suarez$^{5}$, Alin Albu-Schäeffer$^{6}$ and Pedro U. Lima$^{1}$
\thanks{This work is supported by euROBIN, the European Robotics and AI Network (Grant agreement ID: 101070596) funded by the European Commission.}
\thanks{Manuscript received MONTH DAY, 2025; revised MONTH DAY, 2025.}

\thanks{$^{1}$Institute for Systems and Robotics / LARSyS, Instituto Superior Técnico, University of Lisbon}%
\thanks{$^{2}$Laboratory for Automatics and Systems, Instituto Pedro Nunes}
\thanks{$^{3}$INRIA, CNRS, Université de Lorraine, Nancy, France}
\thanks{$^{4}$Munich Institute of Robotics and Machine Intelligence, TUM, Germany}
\thanks{$^{5}$GRVC Robotics Laboratory, USE, Spain}
\thanks{$^{6}$Institute of Robotics and Mechatronics, DLR, Germany}
}



\maketitle
\begin{abstract}
This paper presents a novel framework for cooperative robotics competitions (coopetitions) that promote the transferability and composability of robotics modules, including software, hardware, and data, across heterogeneous robotic systems. The framework is designed to incentivize collaboration between teams through structured task design, shared infrastructure, and a royalty-based scoring system. As a case study, the paper details the implementation and outcomes of the first euROBIN Coopetition, held under the European Robotics and AI Network (euROBIN), which featured fifteen robotic platforms competing across Industrial, Service, and Outdoor domains.

The study highlights the practical challenges of achieving module reuse in real-world scenarios, particularly in terms of integration complexity and system compatibility. It also examines participant performance, integration behavior, and team feedback to assess the effectiveness of the framework. The paper concludes with lessons learned and recommendations for future coopetitions, including improvements to module tracking, scoring mechanisms, and support tools. 
\end{abstract}







\section{Introduction}
\label{sec:introduction}
\IEEEPARstart{B}enchmarking robotic systems remains a long-standing challenge due to the diversity of platforms, tasks, and the absence of broadly accepted standards. As a result, researchers often rely on custom evaluation setups that limit generalizability and slow scientific progress.

Robotics competitions have helped mitigate these issues by defining shared tasks, environments, and evaluation metrics \cite{behnke2006robot,nardi2016robotics,sun2021research,lima2016rockin}. By pushing teams to operate outside controlled laboratory conditions, competitions expose system limitations, promote robustness, and drive technical innovation.

However, most competitions emphasize individual performance and offer few incentives for collaboration. Teams typically work in isolation, leading to duplicated effort, limited interoperability, and delayed knowledge exchange, despite valuable post-event ``lessons learned'' reports.

To address these limitations, this paper introduces the concept of \emph{cooperative competitions}, or \emph{coopetitions}, designed to encourage structured collaboration and increase module transferability. By incentivizing teams to share, reuse, and integrate components (software, datasets, hardware) during the event, coopetitions provide a practical framework for evaluating interoperability and accelerating knowledge transfer. The first euROBIN Coopetition serves as a case study for the framework.


The euROBIN project, Europe’s Network of Excellence on AI-powered Robotics, aims to enable cross-platform transfer of cognition-enabled robotic methods. In line with the Horizon Europe robotics roadmap \cite{euROBIN_SRA2024}, it targets three application domains: (1) robotic manufacturing, (2) personal robots, and (3) outdoor robotics. A key objective is to create infrastructures and incentives to promote reuse of robotic methods across heterogeneous platforms and institutions.

Initial efforts toward collaborative development included the 2023 Robotics Hackathon in Seville, where eight European teams jointly executed an end-to-end parcel delivery scenario involving aerial, outdoor, and domestic robots \cite{suarez2024door}. This event provided an early testbed for cross-platform transferability.

\begin{figure}[t]
    \centering
    \includegraphics[width=\linewidth]{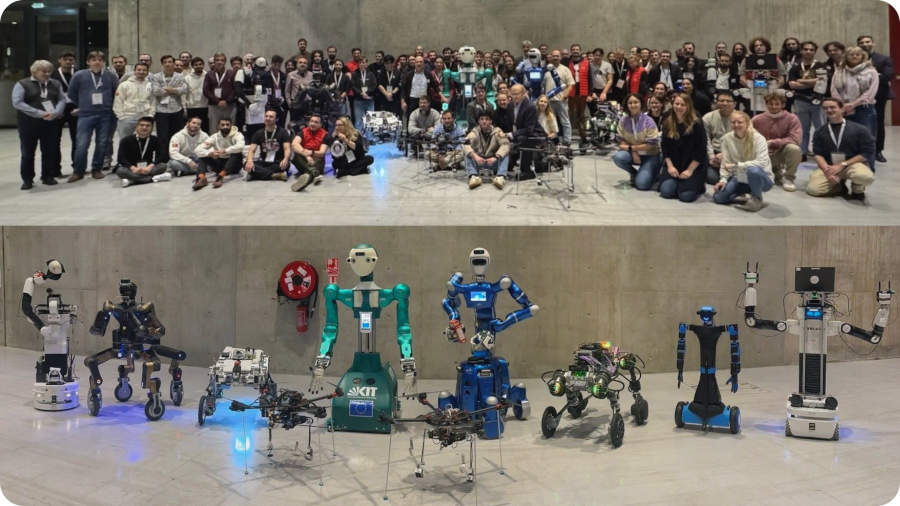}
    \caption{Group photo of the euROBIN Coopetition held in Nancy, France, in November 2024.}
    \label{fig:coopetition_group_photo}
\end{figure}

Building on this foundation, the first euROBIN Coopetition took place in Nancy, France (November 24--28, 2024), bringing together 15 institutions and their robotic platforms across industrial, service, and outdoor domains (Figure~\ref{fig:coopetition_group_photo}). Teams competed while sharing modules and integrating external components to earn transferability bonuses under a royalty-based scoring system.

This paper presents the results of this first coopetition and reflects on its implications for improving reuse of software, data, and hardware modules across heterogeneous robots. The analysis reflects the perspective of the organizing team, aligned with league coordinators and the project coordinator, but may not represent all participant views.

The main contributions of this paper are:
\begin{itemize}
    \item A cooperative competition framework that promotes module sharing and transferability, including a centralized service for distributing and tracking modules.
    \item A set of lessons learned identifying practical and scientific challenges that must be addressed to enable broader reuse of robotic components.
    \item An evaluation of current robotics capabilities in realistic, uncontrolled environments based on observations from the coopetition.
\end{itemize}

For completeness, an online page\footnote{\url{https://eurocore.aass.oru.se:8000/coopetition/results}} provides extended results, milestone descriptions, and additional visualizations. These materials complement but are not required to understand the methods and results presented in this work.
The remainder of the paper is organized as follows: Section~\ref{sec:related_work} summarizes relevant competitions and modularity initiatives, Section~\ref{sec:concept} presents the coopetition framework, Section~\ref{sec:implementation} describes its implementation, Section~\ref{sec:results} reports the outcomes, Section~\ref{sec:lessons_learned} discusses lessons learned, and Section~\ref{sec:conclusion} concludes the work.

\section{Related Work}
\label{sec:related_work}

Robotics competitions have long served as catalysts for advancing the field by providing structured, high-pressure environments that test systems beyond controlled laboratory settings. Events like the DARPA Grand,  Urban, and Subterranean Challenges \cite{seetharaman2006unmanned, buehler2009darpa, tranzatto2022cerberus}, RoboCup \cite{behnke2006robot}, have helped benchmark capabilities in areas such as autonomous navigation, exploration, and mobility. Likewise, RoboCup@Home \cite{matamoros2019trends} and the ANA Avatar XPRIZE \cite{hauser2024analysis} have advanced assistive robotics, human-robot interaction, and telepresence technologies.

The Amazon Picking Challenge \cite{correll2016analysis} and the Mohamed Bin Zayed International Robotics Challenge (MBZIRC) \cite{beul2019team, baca2023autonomous} emphasized perception, manipulation, and multi-robot collaboration, particularly in complex industrial and outdoor scenarios. These events highlighted the increasing importance of heterogeneous teams composed of ground, aerial, and manipulator platforms.

Despite their many benefits, most robotics competitions have prioritized team-specific implementations and competitive performance, offering limited incentives for reusability or collaboration. This often results in duplicated development efforts and limited transferability of components across systems and domains. Initiatives like RoCKIn and the European Robotics League \cite{lima2016rockin} promoted standardized benchmarking but still emphasized direct competition over collaboration.

\begin{figure}[b]
    \centering
    \includegraphics[width=0.9\linewidth]{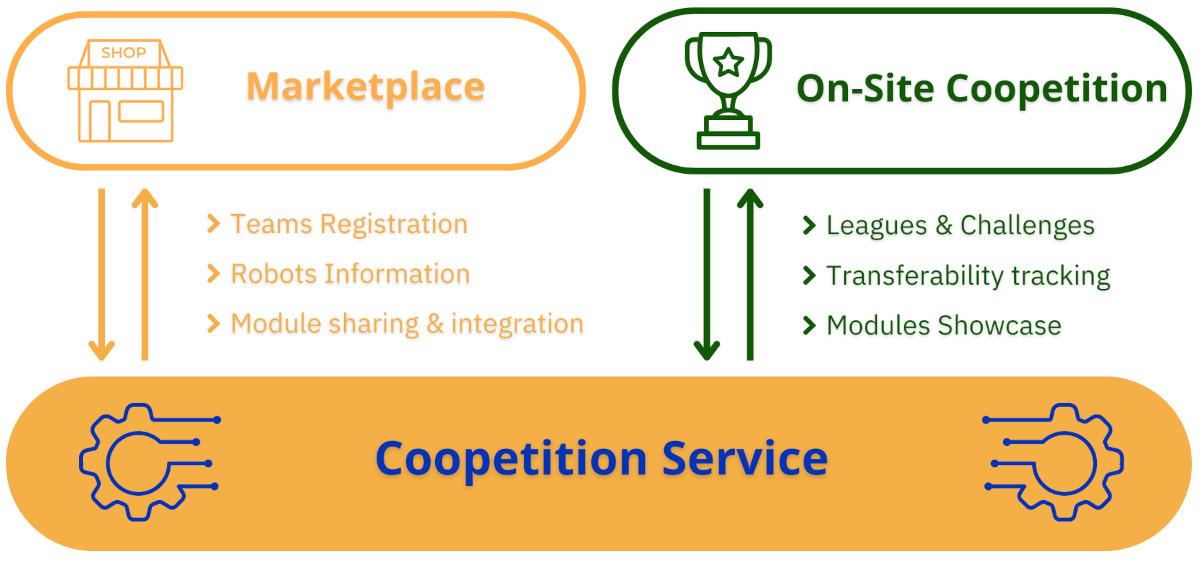}
    \caption{Coopetition design comprised of two stages, and tracked by a coopetition service tailored to its needs.}
    \label{fig:coopetition_structure}
\end{figure}


In parallel, recent years have seen steady progress in modularity and standardization. The ISO 22166 series \cite{zou2022towards} proposes architectures for modular service robots, enabling plug-and-play interoperability at both hardware and software levels. Research efforts like DLR’s modular mechatronics platforms and frameworks such as MoveIt and FlexBE demonstrate how modular design supports reconfiguration and reuse across robotic systems. Middleware like ROS~2 and Open-RMF further facilitate integration in diverse robot teams by offering standardized communication protocols and abstraction layers. 
Complementing these developments, cloud robotics frameworks such as FogROS2-FT \cite{chen2024fogros2} and XBot2D \cite{muratore2023xbot2d} enable computational offloading to edge or cloud infrastructure. These platforms allow high-level modules, such as perception, planning, and mapping, to be reused without requiring on-board deployment, thus promoting robotics-as-a-service paradigms and improving module portability.

Benchmarking efforts have also begun to evaluate generalization and transferability, not just task-specific performance. Competitions such as the Robothon Grand Challenge \cite{so2024digital} introduced Internet-connected shared task board artifacts to provide traceability and automatic scoring to real-world distributed robot performances. These trends collectively support the broader goal of building reusable and interoperable, and shareable robotic software and hardware.

The proposed Coopetition builds on these foundations by operationalizing a new model of cooperative robotics evaluation. It is the first competition designed specifically to incentivize module transferability and collaboration, providing a structured framework to assess integration effort, reuse across teams, and cross-platform applicability of robotics solutions.
\section{Coopetition Framework}
\label{sec:concept}


The Coopetition consists of two stages (Figure~\ref{fig:coopetition_structure}): the Marketplace, which begins months in advance to enable module exchange and early collaboration, and the On-Site Coopetition, where teams solve tasks using their robots. The Coopetition Service supports both stages by tracking module integrations and monitoring scores. More details are provided in the following.

\subsection{Marketplace}
\label{sub_sec:marketplace}
The Marketplace phase is designed to allow teams to add and test modules on their robotic platforms before the On-Site Coopetition begins. During this phase, teams can act as both providers and users. As providers, teams can upload relevant modules, including software, hardware, or data. As users, teams are responsible for tracking the transferability of the modules they acquire, specifying their usage through the Coopetition Service.

This framework adopts a licensing model, in which each module is assigned a royalty value. This value plays a role in the scoring system, as the successful use of a module by another team, specifically when the module is integrated within a robotic platform and used to solve a task, earns both the developer and using team points based on the royalty value associated with the module.

In contrast to a transaction-based approach, where teams buy and sell modules, the licensing model offers greater flexibility. In the former, teams are limited in the number of modules they can test or integrate, which minimizes the potential for transferability. Moreover, a transaction-based approach would require setting initial prices for modules and assigning a budget to each team, which would require pre-defined pricing references. While this framework could be interesting, the proposed framework chose the module licensing approach for its additional flexibility.


Once the On-Site Coopetition phase begins and teams start their task runs, no new modules can be uploaded. This rule is implemented to prevent teams from uploading live task-related data based on their task execution, which can potentially be used by other teams during their own task runs.

While the idea of transferring live data, such as scene graphs or other similar types of information, could be an interesting approach, this type of transferability was considered beyond the scope of the coopetition. This coopetition framework chose to focus on promoting the integration of software and hardware modules rather than enabling live data transfer.

\subsection{On-Site Coopetition}
\label{sub_sec:on_site_coopetition}

Following the Marketplace phase, teams proceed to the Task Demonstration Stage, where they are assigned specific tasks based on their robot type and contextual conditions. Each \textbf{task} is subdivided into multiple \textbf{milestones}, 
and points are awarded for the successful completion of each milestone.

Milestones are deliberately designed to be solvable through multiple strategies, often requiring the integration of several modules. This encourages the transfer of broader capabilities across robotic platforms rather than reliance on narrow, single-purpose solutions. For example, a milestone such as opening a dishwasher may involve repositioning the robot, detecting and localizing the handle, and executing a coordinated motion sequence to open the door.

\textbf{Task difficulty} can vary based on contextual factors. For instance, in the context of domestic service robots, a navigation task may require a robot to move from the kitchen to another room. The complexity of this task depends on whether the path is clear or includes obstacles such as doors or furniture. 

Tasks are also associated with \textbf{task conditional levels}, a mechanism that allows teams to adapt task parameters to better suit their platform capabilities. For example, if the original task specifies navigating from the kitchen to the bedroom, a team may choose to redirect the goal to the living room instead. This adaptation is allowed, but penalized in the final task score, offering teams a trade-off between feasibility and scoring potential.

Each milestone can also be executed under varying \textbf{milestone conditional levels}, which reflect the method used to solve the milestone. Continuing the navigation example, a robot might complete the task fully autonomously, under teleoperation, or by using artificial landmarks to assist localization. These decisions directly affect the milestone score, with higher rewards assigned to more autonomous behaviors.

The scoring system also incorporates \textbf{milestone penalties} to account for undesirable behaviors. For example, if a robot collides with objects while navigating, penalty points are applied to reflect reduced performance in terms of safety and reliability.

Finally, each milestone includes a \textbf{subjective scoring} component, evaluated by external referees, ideally non-experts, who provide qualitative feedback on task execution. This component acts as a "human review," capturing aspects such as fluency, clarity of intent, or user-friendliness that may not be fully reflected in the quantitative metrics.

Bearing these considerations in mind, the execution and effectiveness of each milestone are assessed using the following scoring components:

\begin{equation}
\label{eq:milestone_equation}
MS_{n} = \begin{cases} 
      l_{m_{n}} \left( b_{m_{n}} \left(1 + \frac{q_{m_{n}}}{50} \right) - p_{m_{n}} \right) & \text{if} \;\; (a) \\
      10l_{m_{n}} \left(b_{m_{n}} \left(1 + \frac{q_{m_{n}}}{50} \right) - p_{m_{n}} \right) & \text{if} \;\; (b)\\
      0 & \text{if} \;\; (c)
   \end{cases}
\end{equation}

where (a) $I_{transfer}^{n}=0$, $I_{success}^{n}=1$; 
(b) $I_{transfer}^{n}=1$, $I_{success}^{n}=1$; 
and (c) $I_{success}^{n}=0$.


In Equation~(\ref{eq:milestone_equation}), \(n\) denotes the milestone number, \(b_{m_n} \in \mathbb{N}\) is the base score for milestone \(n\), \(l_{m_n} \in [0,1]\) relates to the milestone conditional level, and \(q_{m_n} \in [0,10]\) is the milestone subjective sub-score that can increase \(b_{m_n}\) by up to 20\%. The indicator function \(I_{transfer}^n : M \rightarrow \{0,1\}\) is 1 when a transferred module is used, and 0 otherwise. The term \(p_{m_n} \in \mathbb{N}\) accounts for penalties incurred during milestone execution, while \(I_{success}^n : M \rightarrow \{0,1\}\) is 1 if the milestone was successfully completed.

Equation~(\ref{eq:milestone_equation}) defines how points are awarded for each milestone. A team receives zero points if the milestone is not completed $(c)$, the baseline score if it is successfully completed $(a)$, and a score equal to ten times the baseline if the milestone is completed using a module from another team $(b)$.

The values for $b_{m_{n}}$ (baseline score), $l_{m_{n}}$ (milestone conditional level), and $p_{m_{n}}$ (penalty) are defined by the coopetition organizers and referees prior to the start of the On-Site Coopetition. During execution, $I_{success}^{n}$ (milestone success indicator) and $q_{m_{n}}$ (subjective score) are determined by referees and external evaluators based on team performance. In contrast, $I_{transfer}^{n}$ (transferability indicator) is automatically managed by the Coopetition Service, based on team declarations and system tracking. Teams are responsible for informing the referees of their selected task and milestone conditional levels before executing each run.

The final score for a task can be computed as follows:

\begin{equation}
\label{eq:task_equation}
S_{task} = T \sum_{n=1}^{N} \Big(1 - I_{transfer}^{n}\sum_{k=1}^{M_n}\frac{r_{(n,k)}}{M_{n}} \Big)\max\{0, MS_n\}
\end{equation}


where \(MS_n\) is the score of milestone \(n\), \(N \in \mathbb{N}\) denotes the total number of milestones in the task, and \(M_n\) is the number of external modules integrated to solve milestone \(n\). The index \(k\) refers to a specific module, with \(r_{(n,k)}\) representing its associated royalty fee. The factor \(T\) corresponds to the selected task conditional level and scales the overall task score.

In Equation~(\ref{eq:task_equation}), $I_{transfer}^{n}$ is a binary variable (0 or 1) that determines whether the royalty value impacts the scoring system. If $I_{transfer}^{n} = 1$, the score obtained by the user team for a given milestone is adjusted according to the royalty share, while the provider team receives the corresponding fraction. Specifically, the user team retains a portion of the milestone score equal to $1 - r$, where $r$ is the normalized royalty value. For each milestone, the royalties associated with all integrated modules are summed and then divided by the total number of modules used, thereby normalizing the royalty distribution across multiple integrations.

The sum of all task scores is computed as:

\begin{equation}
\label{eq:execution_equation}
S_{challenge} = \sum_{i=1}^{S}S_{task}^i
\end{equation}

where ${i}$ is the task number, ${S} \in \mathbb{N}$ is the total number of tasks, ${S_{task}^i}$ is the score for task $i$, and $S_{challenge}$ represents the teams score on a Challenge.



Equation~(\ref{eq:task_equation}) defines the task score for a module user team. In the case of transferability, the share of the corresponding developer team(s) is calculated according to the following metric:

\begin{equation}
\label{eq:task_equation_2}
R_{task} = \sum_{n=1}^{N} I_{transfer, {t}}^n \frac{1}{M_{n}} \sum_{k=1}^{{M_{t_d}}} \frac{r_{t_d, (n,k)}}{T_k} MS_{(n, t)}
\end{equation}


where, \(t_d\) denotes a developer team, while \(t \neq t_d\) refers to a team that integrated at least one module developed by \(t_d\), since teams cannot earn royalties from their own modules. The term \(I_{transfer,t}^n\) indicates the transfer indicator for module(s) developed by \(t_d\) and used by team \(t\) in milestone \(n\). The variable \(M_{t_d}\) represents the number of modules developed by \(t_d\) and integrated by team \(t\) to solve milestone \(n\), with \(r_{t_d,(n,k)}\) denoting the royalty percentage assigned to module \(k\) in that milestone. The term \(T_k\) accounts for the number of teams that jointly developed module \(k\) with \(t_d\). Finally, \(M_n\) is the total number of modules integrated by team \(t\) for milestone \(n\), and \(MS_{(n,t)}\) corresponds to the milestone \(n\) score achieved by team \(t\).

\begin{figure*}[t]
    \centering
    \includegraphics[width=0.75\textwidth]{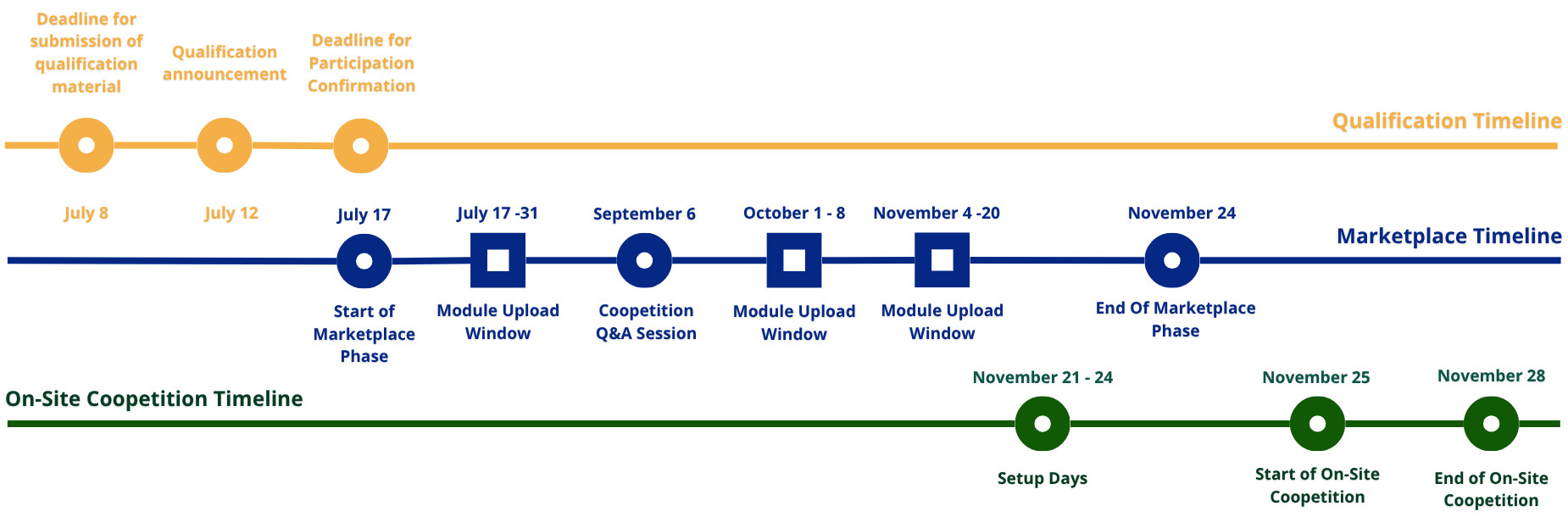}
    \caption{Coopetition timeline.}
    \label{fig:coopetition_timeline}
\end{figure*}

Equation~(\ref{eq:task_equation_2}) accounts not only for the total number of integrated modules used in a milestone but also for the total number of contributing developer teams, denoted as $T_k$. The intuition behind it is similar to Equation~(\ref{eq:task_equation}), where normalization ensures a bound distribution of rewards, this time extending the normalization across all contributing developers.

Since modules can be reused across different leagues, teams, and tasks, the final royalty score for a developer team is computed as:

\begin{equation}
\label{eq:task_equation_3}
S_{royalties} = \sum_{l=1}^{L} \sum_{\substack{t=1 \\ t \neq t_d}}^{T} \sum_{i=1}^{S} R_{task}^{(l, t, i)} 
\end{equation}

where ${R_{task}^{(l, t, i)}}$ represents the royalties score for task ${i}$ of league ${l}$ taken from team ${t}$, ${L}$ represents the total number of leagues, ${T}$ represents the total number of teams, and $S_{royalties}$ represents the points earned from licensing modules across leagues, teams, and tasks.


The team's final score is the sum of the Challenge score and the points earned through module transferability, expressed as \( S_{\text{coopetition}} = S_{\text{challenge}} + S_{\text{royalties}} \).

The current scoring system assumes that when a team uses multiple modules from different providers, the royalty share is evenly distributed based on the number of modules and the number of developer teams involved.

From the user team's perspective, integrating one or several modules, regardless of whether they originate from the same or different teams, yields the same milestone score. Moreover, the system does not differentiate between module types or consider the complexity of their integration, meaning no additional points are awarded for incorporating more technically demanding modules.



\subsection{Coopetition Service}
\label{sub_sec:coopetition_service}

The Coopetition Service was developed as an online platform to support collaboration and module transferability among participating teams. It enabled teams to share software, hardware, and data modules, describe their robots and systems, and track their performance throughout the competition. The service featured dedicated sections for managing team and robot profiles, publishing and discovering modules via a “Marketplace,” and manually declaring the integration of external modules to gain transferability points.
Teams could monitor their progress, integration benefits, and receive notifications on module usage. Referees had additional privileges for task creation, milestone definition, and score management, while also verifying module usage during task execution. The system was implemented using Nuxt.js for the web framework and Prisma as the object-relational mapper for database management. 

\textit{The Coopetition Service is accessible at}
\href{https://}{this link}.

\section{Coopetition Implementation}
\label{sec:implementation}

The coopetition framework was implemented during the inaugural euROBIN Coopetition, 
following the timeline shown in Figure~\ref{fig:coopetition_timeline}. 
The Marketplace phase ran from July 17 until the day before the On-Site event and 
included three module upload windows, a Q\&A session, and access to a Rulebook 
detailing procedures, scoring, and league-specific guidelines.
\textit{The official Rulebook is available at} \href{https://}{this link}.
Before the Marketplace, teams inside and outside euROBIN were invited to submit qualification materials. These consisted of a team description paper outlining achievements, research focus, and real-world applicability, and a video demonstrating modules or robot capabilities. External teams were required to showcase tasks aligned with their target league, while euROBIN teams were automatically qualified.

\begin{figure}[t]
    \centering
    \includegraphics[width=1\linewidth]{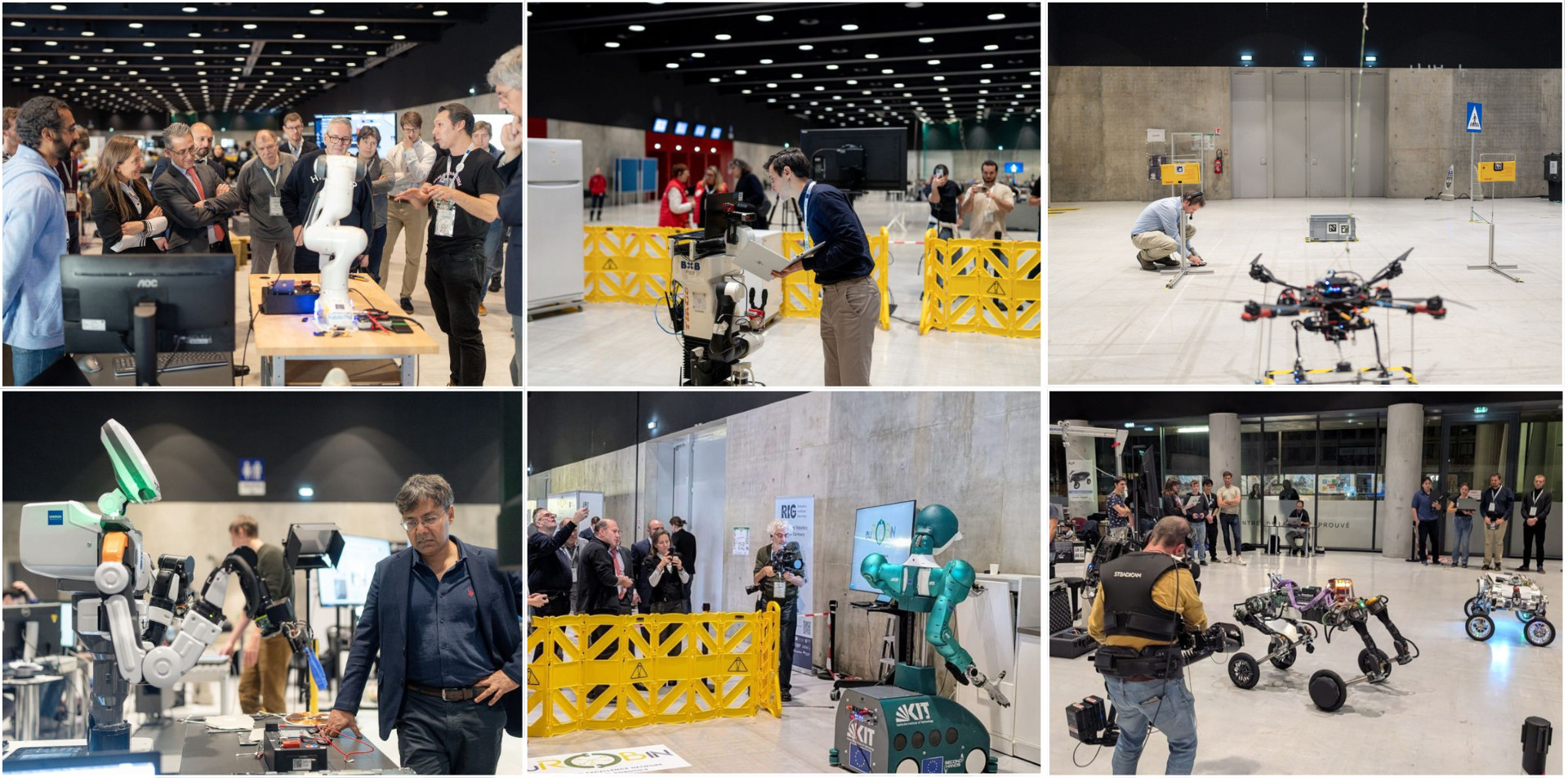}
    \caption{Teams participating in the euROBIN Coopetition across the Industrial (left), Service (center), and Outdoor Robots Leagues (right images). Image credits: © Inria / Photos B. Fourrier and Justine Galet.}
    \label{fig:league_teams_photo}
\end{figure}

The coopetition featured three leagues, Industrial Robots (IRL), Service Robots (SRL), and Outdoor Robots (ORL) shown in Figure~\ref{fig:league_teams_photo}. These focused respectively on manipulation, assistive and domestic robotics, and autonomous navigation in dynamic environments. Although all leagues followed a shared structure and general rules, each defined its own tasks and milestones. 


Scoring was league-specific, with one winner per league. A fixed royalty rate of $r_n = 0.25$ allowed module developers to earn up to 25\% of milestone points when their modules were reused. This value was chosen to ensure stability during the first edition, given limited upload windows, no prior experience, and an initially small marketplace.

Each league had a dedicated referee team: an organizer, an expert referee, and a non-expert referee. Organizers and experts assessed milestone success, conditional levels, and penalties, while non-experts provided subjective assessments. A technical committee oversaw module usage claims through the Coopetition Service and could remove non-compliant or redundant modules from the Marketplace.

The following sections describe each league in detail.

\subsection{Industrial Robots League (IRL)}
\label{sub_sec:irl}

The IRL focused on robotic manipulation and was inspired by the euROBIN Manipulation Skill Versatility Challenge \cite{so2024eurobin}. Its main event, the \textit{Task Board Manipulation Challenge}, introduced an industry-supported benchmark using an internet-connected electronic task board. 


\begin{table}[b]
\centering
\scriptsize  
\setlength{\tabcolsep}{4pt}  
\begin{tabular}{c c p{4.2cm}}
\hline
\rowcolor[HTML]{b96e64}
\textbf{Team} & \textbf{Institution} & \textbf{Robot / End Effector / Sensors} \\
\hline
\rowcolor{rowred}
TUM-MIRMI & TUM & Franka FR3 / Franka Hand / Realsense D435i \\
HCR Team & J. Stefan Inst. & Franka FR3 / Franka Hand / Realsense D435 \\
\rowcolor{rowred}
Tecnalia Flexbotics & Tecnalia & Nextage / Pneumatic EE / Stereo Vision \\
Fraunhofer IPA & Fraunhofer IPA & UR5e / WSG 50-110 / Realsense D435 \\
\rowcolor{rowred}
OSCAR & CEA & UR10e / Robotiq 2F-140 / Realsense D435, Phoxi M, FT Sensor \\
\end{tabular}
\caption{IRL teams, institutions, and robots.}
\vspace{-15pt}
\label{tab:industrial_robots_teams}
\end{table}

Participants developed fully automated solutions to interact with the board, which was randomly placed on a worktable. Robots first located the board and then executed a predefined sequence of actions simulating typical industrial electronics cabinet operations, all within a fixed time limit. Manipulation skills assessed included object localization, interactive manipulation, insertion, levering, wrapping, and pressing. 

Points were awarded for the successful completion of specific milestones, which were tracked through performance feedback from the task board circuitry.

Each trial began with referees resetting the board to its nest position and placing it randomly. The timer started when the team leader pressed the start button and ended when the robot pressed stop or the 10-minute limit was reached. Teams could opt for teleoperation, in which case referees specified the operator’s location. Table~\ref{tab:industrial_robots_teams} lists the participating teams, their institutions, and robotic platforms. 

\subsection{Service Robots League (SRL)}
The SRL focused on advancing service and assistive robots for personal and domestic settings. It featured the \textit{Multi-Functional Service Robot Challenge}, where teams received a randomly selected command from a predefined set, forcing real-time planning rather than scripted actions.

The challenge comprised three tasks of increasing difficulty:

\begin{enumerate}
  \item Pick-and-place/give within the selected \textbf{base} kitchen.
  \item Pick actions in the \textbf{base} kitchen and place/give actions in a kitchen randomly selected by referees.
  \item Both pick and place/give actions in kitchens randomly selected by referees.
\end{enumerate}

Two execution scenarios were possible: completing the task entirely within one kitchen or retrieving an object from one kitchen and delivering it elsewhere or to a person. Teams designated a \textbf{base} kitchen, the environment where they felt most confident, at the start of the event and informed referees before each task.

This challenge tested speech recognition, navigation, object/person detection, object feature recognition, manipulation, and task-level planning. Commands were dynamically generated at runtime using a command generator supervised by referees, which is publicly available at
\href{https://}{the project repository}. Possible task elements, such as kitchens, locations, actions, and objects, followed predefined sets (Table~\ref{tab:srl_combined}). Example commands included: “Pick the \textbf{Pringles} from the \textbf{INRIA cabinet} and give them to the person in the \textbf{KIT kitchen}.”


To mitigate unpredictability, teams could specify preferred actions, locations, or objects (\textbf{task conditional levels}). 

Milestones were categorized by primary action (Navigation, Speech, Manipulation, or Perception) to simplify scoring in the coopetition service. Although milestones often required multiple actions, this categorization emphasized their main objective. Teams could also choose to manipulate unknown objects supplied by organizers, without prior exposure, or known objects selected from YCB and KIT dataset subsets.

Each team had up to three 10-minute attempts per task, with the best attempt used for scoring. Tasks began after arena setup and conditional level selection, followed by a referee-issued command.

\begin{table}[t]
\centering
\scriptsize
\setlength{\tabcolsep}{5pt}
\begin{tabular}{c c c}
\hline
\rowcolor{headergreen}
\textbf{Team} & \textbf{Institution} & \textbf{Robot} \\
\rowcolor{rowgreen}
Alter-Ego   & University of Pisa        & Alter-Ego \\
INRIA       & INRIA                     & Inria Tiago Dual \\
\rowcolor{rowgreen}
SUSR Team   & ISIR, Sorbonne Université & TIAGo \\
DLR         & DLR                       & Rollin' Justin \\
\rowcolor{rowgreen}
KIT         & H2T                       & ARMAR-7 \\
SocRob@Home & ISR (IST)                 & TIAGo \\
\rowcolor{rowgreen}
GEPETTO     & LAAS, CNRS                & {} \\
\end{tabular}

\vspace{6pt} 

\begin{tabular}{ c c c}
\hline
\rowcolor{headergreen}
\textbf{Kitchen ($K_i$)} & \textbf{Location ($L_i$)} & \textbf{Object ($O_i$)} \\
DLR   & Dishwasher & YCB Dataset Subset \\
\rowcolor{rowgreen}
KIT   & Table      & KIT Dataset Subset \\
INRIA & Cabinet    & Unknown Objects \\
\rowcolor{rowgreen}
{}    & Counter    & {} \\
\end{tabular}

\caption{Teams and robot platforms participating in the SRL (top), and task-relevant kitchen, location, and object constraints for the Multi-Functional Service Robot Challenge (bottom).}
\vspace{-15pt}
\label{tab:srl_combined}
\end{table}




\subsection{Outdoor Robots League (ORL)}
The ORL featured the \textit{Delivery Robot Challenge}, designed to advance autonomous delivery capabilities in both aerial and ground-based robots.

Teams were tasked with retrieving and delivering parcels across dynamic indoor and semi-outdoor environments through three increasingly difficult tasks:

\begin{enumerate}
\item \textbf{Ground robots:} Easy route within the venue, medium route to the entrance hall, and hard route through the logistics area/emergency exit.
\item \textbf{Aerial robots:} Easy route with no obstacles, medium route with small/medium obstacles, and hard route with large obstacles and narrow passages.
\end{enumerate}

In all scenarios, robots collected a parcel from a designated pick-up location and delivered it to a specified point. Core skills evaluated included navigation in complex environments, manipulation, task planning, object detection, and recognition.

UGV route difficulty increased with each task. Task 2 required door crossings and stair navigation, while Task 3 added ramps, closed doors through logistics areas. Dynamic obstacles such as people, chairs, cones, CrossFit boxes, and large containers were added along the way. The UAV, operating in the flying area covered by a safety net, faced obstacles such as shelves, cones, and boxes.

\begin{table}[b]
\centering
\vspace{5pt}  
\begin{tabular}{m{1cm} m{2cm} m{4cm}}
\hline
\rowcolor{headerblue}
\textbf{Team} & \textbf{Institution} & \textbf{Robot} \\
\rowcolor{rowblue}
USE   & GRVC Robotics Lab        &  Dual Arm Aerial Manipulation Robot \& LiCAS A1\\
RSL   & RSL ETH Zurich & Transformer Robot \& Leva \\
\rowcolor{rowblue}
IIT       & Istituto Italiano Di Tecnologia        & Centauro \\
\end{tabular}
\caption{ORL teams, institutions, and robots.}
\label{tab:orl_robots_teams}
\end{table}

Task parameters (e.g., actions, pick-up and delivery points, parcels) were assigned at runtime using a command generator supervised by referees. Examples include: "Pick parcel \textbf{A2} from Pick-Up \textbf{Point A} and deliver it to \textbf{Point J}".

Parcels included two small drone-compatible boxes (IDs A0 and A1) and one larger UGV box (ID A2), all tagged with ArUco markers. 

Teams could use \textbf{task conditional levels} to adapt tasks to their platforms. Each team had three 10-minute attempts per task, with the best score counting as the final result. Tasks began once robots were positioned, conditional levels declared, and the arena prepared. Table~\ref{tab:orl_robots_teams} shows the participating teams.

\section{Coopetition Results}
\label{sec:results}
This section reports on the outcomes of the first euROBIN Coopetition, beginning with module sharing and integration patterns observed during the Marketplace and On-Site phases. Then it concludes with insights gathered from the post-event questionnaire.


\subsection{Module Sharing \& Integration}
\label{sec:results_marketplace}
During the Marketplace phase, three Module Upload Windows took place: 24 modules were uploaded in the first window, 32 in the second, and 34 in the third, totaling 90 modules available for integration.

\begin{figure*}[t]
    \centering
    \includegraphics[width=0.85\linewidth]{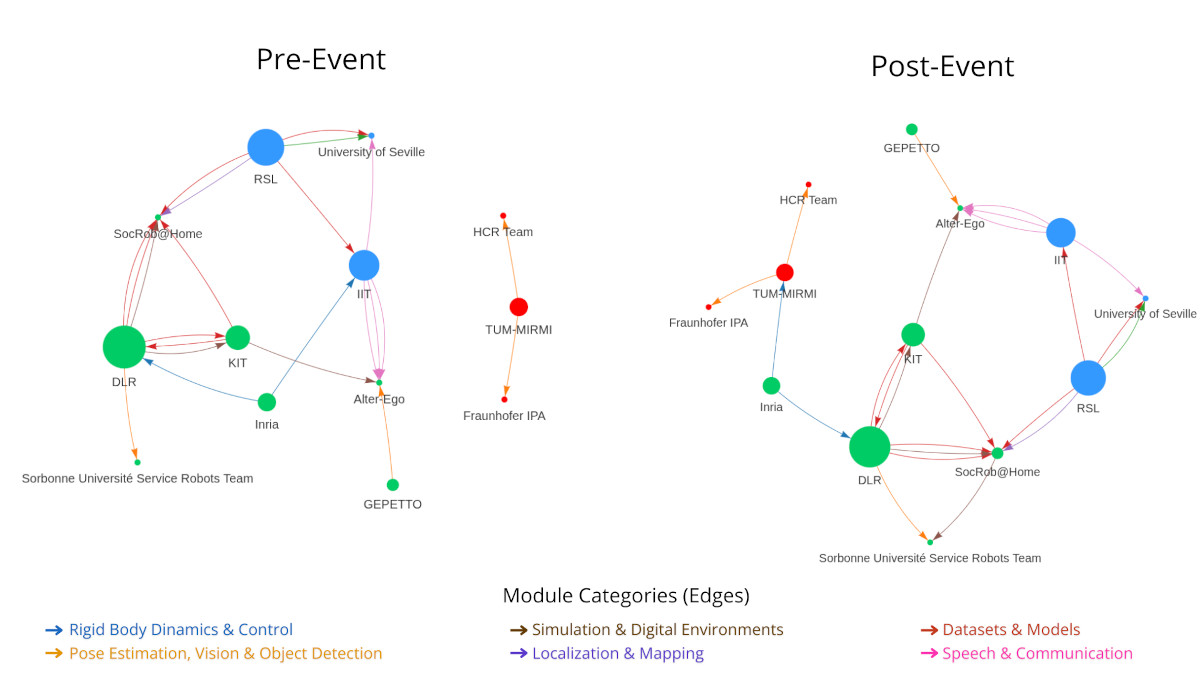}
    \caption{Module Transferability Social Graphs during the Marketplace (Pre-Event) and On-Site Coopetition (Post-Event) phases. The nodes represent the teams, and the edges indicate the direction of module integration. For example, “INRIA → IIT” indicates that INRIA developed a module, which was then integrated by IIT. The node sizes are proportional to the amount of royalties the team receives, with larger nodes indicating more royalties. The node color corresponds to the league the team participated in. The edge color represents the general category of the modules as illustrated in the image.}
    \label{fig:module-analysis}
\end{figure*}

Figure~\ref{fig:module-analysis} illustrates module transferability, i.e. module availability and integration, during the Marketplace (Pre-Event) and On-Site Coopetition (Post-Event) phases. 
In the pre-event and post-event networks, we can observe a noticeable increase in module integration during the On-Site Coopetition phase, as new edges were added to the graph. Initially, two separate social graphs were formed, i.e. the IRL was isolated from the other leagues. By the end of the event, these graphs merged into a single connected graph that included all the leagues. Modules linked to these edges included:


\begin{itemize}
    \item \textbf{Rigid Body Dynamics \& Control}: Pinocchio Library, OpenSoT \& CartesI/O
    \item \textbf{Pose Estimation, Vision \& Object Detection}: TUM Task Board Perception Package, HappyPose, 3D Object Tracking, YOLO ROS
    \item \textbf{Simulation \& Digital Environments}: BlenderProc, Nancy Digital Twin, Kitchens Envs, DLR Kitchen
    \item \textbf{Localization \& Mapping}: ICP Localization
    \item \textbf{Datasets \& Models}: KIT Object Dataset, Table obj
    \item \textbf{Speech \& Communication}: Speech-To-Text, Text-To-Speech, LLM Task Planning
\end{itemize}






Upon analyzing the distribution of module types and their integration patterns across the different teams and leagues, it becomes evident that teams preferred sharing data modules over software modules. Data modules, such as furniture CAD models and maps, were made available by the teams and used across multiple leagues. In line with this trend, modules categorized as Pose Estimation, Vision \& Object Detection, and Speech \& Communication were also frequently used.

As expected, integrating high-level modules like perception algorithms or large language models (LLMs) into a pipeline is generally easier than integrating low-level modules, such as controllers, which are often platform-dependent. Since the scoring system does not evaluate the integration effort based on module types, teams are likely to opt for what is perceived as the "easiest" solution, either by using available data or integrating high-level modules for perception, speech, and planning.

While this was the general trend, some teams made an effort to integrate more complex modules, such as Rigid Body Dynamics and Control algorithms, including the Pinocchio Library and OpenSoT \& CartesI/O.


Across all leagues, three key trends in module transferability emerged. First, perception and communication modules saw the highest levels of reuse, primarily due to their platform-agnostic nature. In contrast, low-level or hardware-interfacing modules were rarely reused, as their integration required significant effort and was often constrained by platform dependencies. Finally, more complex tasks tended to encourage module reuse, but only when the integration workload remained feasible for participating teams.




\subsection{Teams Questionnaire \& Feedback}
\label{sec:results_Questionnaire}


Following the On-Site Coopetition, a questionnaire\footnote{\url{External link removed for double-anonymous review}} was distributed to gather feedback on modules, transferability, and coopetition design. A total of 16 responses were collected, with at least one submission from each participating team.

\subsubsection{Modules, Standardization, and Transferability}
\label{sec:results_standardization}
The definition of what constitutes a “module” remained unclear, with varying interpretations ranging from low-level controllers to high-level systems.

Opinions on standardization were similarly divided. Some supported it, others were unsure, and some opposed it. Concerns among those against standardization included its potential to slow innovation, especially given the current maturity level of robotics and AI technologies. 

Supporters of standardization recommended starting with conventions for data formats, communication protocols, and reproducible environments. They proposed aligning message interface definitions, unifying robot and scene description formats, and leveraging existing tools such as Docker, ROS~2, and YAML, while acknowledging the technical limitations that still need to be addressed. 
Some also recommended community-driven processes for evolving standards, such as proposal-based models. In contrast, others preferred minimalistic approaches, relying on simple repositories with clear documentation and flexible integration practices.

Integration strategies varied, with most teams lacking a formal process. Some used Docker or Singularity for isolation and ROS for communication.

Regarding module usage, 31.3\% of teams developed only their own modules, 37.5\% used third-party modules, and the remaining 31.3\% did both. Challenges to integrating third-party modules included interface incompatibilities, high integration effort, platform dependencies, lack of documentation, non-reproducible builds, and poorly defined inputs/outputs. These issues often led teams to reimplement modules to better understand or adapt them.

From the coopetition perspective, limited time and difficulty in module integration were commonly cited. Some teams received support from other developers, while others did not, often due to the lack of availability during the On-Site phase.


\subsubsection{Coopetition Design}
\label{sec:results_coopetition_design}

Teams provided feedback on various aspects of the coopetition's design, including module assessment, scoring, and event structure. Opinions varied on whether the fixed royalty value assigned to all modules fairly reflected the effort required for their integration. Some teams suggested that more complex modules should warrant higher rewards, while others proposed allowing negotiation of royalty values. There was also interest in incentivizing multiple integrations.

Most teams reported understanding the scoring system and its transferability rewards, though a few indicated confusion about certain parameters. Survey results showed high appreciation for task and milestone conditional levels, while penalties and subjective scoring were seen as moderately useful. 

The Module Upload Windows received mixed responses. Some teams found them helpful for planning, while others felt restricted by the inability to add new modules during the On-Site phase. Suggestions included holding pre-event gatherings to facilitate earlier collaboration.

On module tracking, most teams supported replacing the trust-based module tracking with a system that requires proof of integration, potentially verified by developers. This would likely require more time per team during task execution.


In terms of future design, teams proposed cooperative or cross-league tasks, environment transferability, and formats encouraging collaboration, such as mixed-team challenges or group competition against a human baseline. Several teams also suggested keeping current task difficulty while enhancing the design and retry rules, and emphasized the need for more pre-event coordination and visibility for shared modules.

Some concerns were raised over scoring that favored teleoperation and about the tight scheduling, with teams advocating for more time between attempts and greater focus on full task autonomy.

\section{Lessons Learned}
\label{sec:lessons_learned}
This section summarizes the main lessons from the coopetition, focusing on the effectiveness of the framework in promoting transferability and reflecting on team performance.


\subsection{Coopetition Framework \& Transferability}
\label{sec:framework_lessons}
The coopetition framework proved effective in promoting collaboration and module reuse, even without a clear, unified definition of what constitutes a “module”. Despite differing research areas, teams were able to integrate and adapt each other’s solutions, often relying on external modules to cover functionalities outside their core expertise. This need-driven transferability helped reduce redundant effort across the event.

However, the robotics ecosystem remains far from “plug-and-play”. Alignment in message interfaces, robot and scene description formats, and module input/output specifications would help lower integration barriers. Teams also emphasized the importance of clearer documentation and accessible tutorials to improve module usability.

Module sharing was another limiting factor. Although the coopetition incentivized open-source contributions, several participants especially those using proprietary code, were unable to share certain modules. As industry involvement grows, future editions should consider mechanisms such as selective sharing agreements or alternative contribution models to balance openly released contributions with intellectual property constraints.

The scoring system successfully encouraged transferability through module usage bonuses, but several aspects require adjustments. The fixed royalty structure assumes that software, data, and hardware modules contribute equally, overlooking significant differences in integration effort. A more nuanced system, focused on rewarding complex, low-level modules more than easily integrable, high-level ones, would better reflect real integration costs. Likewise, the current mechanism does not incentivize integrating multiple modules, as the bonus remains unchanged regardless of the number of integrations.

Module tracking also requires improvement. The existing trust-based verification is difficult to audit. An automated logging mechanism would enhance fairness, transparency, and reproducibility. Future coopetitions should collect quantitative measures of integration effort (e.g., integration time or teams feedback) to better understand module complexity and integration cost.

Finally, while conditional levels allowed teleoperation, the scoring sometimes may have favored teleoperated over autonomous behaviors, misaligning incentives with long-term research goals. Clearer differentiation, such as having a separate league or scoring, for autonomous and teleoperated execution would help address this issue.

To strengthen cross-team collaboration, future editions should introduce explicitly cooperative tasks and provide structured opportunities, such as pre-event workshops or integration sessions, for teams to discuss solutions and explore synergies. Participants consistently reported that limited time during the On-Site phase hindered deeper collaboration.

\subsection{Teams \& Robots Performances}
\label{sec:lessons_performances}

The euROBIN Coopetition offered a broad view of current capabilities in European robotics, highlighting progress in perception and navigation while underscoring persistent challenges in manipulation, generalization, and operation in unstructured environments.

In the IRL, teams demonstrated strong manipulation skills in tasks such as circuit probing and cable wrapping. However, performance declined in later milestones requiring dynamic perception and high-precision actions. Although subjective scores were high for most completed milestones, referees noted that prior access to the task board may have encouraged tailored, task-specific solutions, raising concerns about robustness under more randomized conditions.

In the SRL, teams struggled primarily with manipulation and generalization across environments. Even in their preferred base kitchens, robots faced difficulties with cabinet, drawer, and dishwasher manipulation due to complex motion planning requirements. These challenges intensified in Tasks~\#2 and \#3, which introduced randomly assigned kitchens. Navigation and speech-based command understanding were generally reliable, but success rates dropped when objects were occluded, unknown, or surrounded by clutter. Fully autonomous manipulation remained rare, and several teams relied on teleoperation or human assistance. Nonetheless, this league exhibited the highest degree of module reuse, with many successful milestones where some integrated modules led to improved outcomes.

In the ORL, results must be interpreted with caution due to the small sample: one UAV and two UGV teams, one of which suffered major hardware failures. UGVs struggled most with manipulation and command interpretation, often relying on teleoperation to handle parcels. Performance dropped further in Task~\#2, recovering only partially in Task~\#3 despite the use of task conditional levels to simplify routes. Dynamic navigation on staircases or crowded areas remained a major barrier. By contrast, the UAV team showed consistently stronger performance across tasks, successfully completing most deliveries. However, safety constraints in the flight area and obstacle density limited the ability to assess performance in more demanding scenarios.

Overall, results across the three leagues confirm that autonomy in previously unknown, dynamic, and unstructured settings remains an open challenge, particularly for manipulation tasks.


\section{Conclusion \& Future Work}
\label{sec:conclusion}

This paper presented a cooperative coopetition framework to promote module transferability across teams. The first euROBIN Coopetition served as a concrete case study, revealing both promising results and important limitations in module integration, task execution, and cross-team collaboration.

The event showed module reuse can reduce development effort and, in some cases, improve task performance. Yet, low-level and hardware-dependent modules remained difficult to integrate due to platform specificity. While opinions diverged on enforcing standards, many teams agreed that shared data formats and communication protocols would ease integrations.

Open released contributions are also a current challenge. Some teams could not share modules due to licensing constraints. As industrial participation increases, mechanisms such as selective-sharing agreements may be needed to support both academic and commercial contributors.

The scoring system incentivized transferability but treated all module types equally, rewarding simple data modules as much as complex controllers. Future editions should differentiate module types and consider integration effort, supported by metrics such as integration time or compatibility issues.

Likewise, the trust-based module-tracking process limited verification, and automated logging would improve transparency. Distinguishing autonomous from teleoperated execution would be also essential, as teleoperation affected comparability across teams.

Moreover, the event revealed opportunities for deeper collaboration. Limited On-Site time constrained cross-team integration, suggesting the need of pre-event workshops and cooperative or cross-league tasks.

From a technical standpoint, the coopetition showed persistent challenges in manipulation, generalization, and environment transferability, even in partially familiar settings.

Overall, the coopetition functioned both as a benchmark and as a living laboratory for studying collaboration, and integration challenges. It offers a replicable foundation for future community-driven efforts to advance transferability in robotics.

\section*{Acknowledgments}
The authors thank the Principal Investigators and team leaders of the groups that participated in the coopetition case study for their valuable contributions and for providing the data used in this work:
Ales Ude (Jozef Stefan Institute), 
Francisco Blanco (Tecnalia),
Daniel Bargmann and Werner Kraus (Fraunhofer IPA),
Caroline Vienne and Lucas Labarussiat (CEA),
Tamim Asfour (KIT),
Serena Ivaldi (INRIA),
Eleonora Sguerri, Manuel G. Catalano, and Antonio Bicchi (University of Pisa),
Stéphane Doncieux (Sorbonne Université),
Olivier Stasse (LAAS-CNRS),
Anibal Ollero (USE),
Marco Hutter (ETHZ),
and Nikos Tsagarakis (IIT).

\printbibliography


\end{document}